\documentclass[11pt]{article}
\usepackage[hyperref]{ccl2025-en}
\usepackage{times}
\usepackage{url}
\usepackage{latexsym}
\usepackage{fancyhdr}

\usepackage{graphicx}
\usepackage{amsmath}
\usepackage{amssymb}
\usepackage{tabularx}
\usepackage{makecell}
\usepackage{multirow}
\usepackage{booktabs}

\title{DualReward: A Dynamic Reinforcement Learning Framework for \\Cloze Tests Distractor Generation}

\author{Tianyou Huang\textsuperscript{1,a}, Xinglu Chen\textsuperscript{1,a}, Jingshen Zhang\textsuperscript{1}, Xinying Qiu\textsuperscript{1,*}, Ruiying Niu\textsuperscript{2}\\
	\textsuperscript{1}Department of Computer Science, School of Information Science and Technology, \\
	\textsuperscript{2}Faculty of English Language and Culture\\
	Guangdong University of Foreign Studies, Guangzhou, China \\
	{\tt tenny314159@gmail.com, luc88749@gmail.com, audbut0702@163.com, } \\
	{\tt xy.qiu@foxmail.com, 199610539@gdufs.edu.cn} \\}

\date{}

\begin{document}
\maketitle
\begin{abstract}
  This paper introduces DualReward, a novel reinforcement learning framework for automatic distractor generation in cloze tests. Unlike conventional approaches that rely primarily on supervised learning or static generative models, our method employs a dual reward structure with adaptive scaling that differentiates between human-created gold standard distractors and model-generated candidates. The framework dynamically adjusts reward signal intensity based on model performance and confidence. We evaluate our approach on both passage-level (CLOTH-F) and sentence-level (MCQ) cloze test datasets, demonstrating consistent improvements over state-of-the-art baselines. Experimental results show that our adaptive reward scaling mechanism provides modest but consistent benefits on homogeneous datasets (CLOTH-F) and more substantial improvements (3.48-3.86\% in P@1) on diverse, cross-domain data (MCQ), suggesting its particular effectiveness for handling varied question types and domains. Our work offers a flexible framework that effectively balances learning from reliable human examples while exploring novel, high-quality distractors for automated test generation.
  \englishkeywords{Distractor Generation \and Cloze Test \and Reinforcement Learning}
\end{abstract}

\section{Introduction}
\label{intro}

%
%
\cclfootnote{
    %
    %
    \textsuperscript{a}Equal Contribution; \textsuperscript{*}Corresponding author\\
    \hspace{-0.65cm}  
    \textcopyright 2025 China National Conference on Computational Linguistics

    \noindent Published under Creative Commons Attribution 4.0 International License
}

\label{intro}
Educational assessments play a critical role in evaluating student knowledge and comprehension, with cloze tests (fill-in-the-blank questions) being widely used due to their effectiveness in measuring language proficiency and content understanding. Automatic distractor generation (ADG) has emerged as a promising research direction to address the challenge of creating effective educational assessments, with diverse approaches evolving rapidly in recent years  \cite{alhazmi2024distractor}. Earlier methods in this field employed knowledge-driven approaches that leverage semantic hierarchies and taxonomies to identify plausible distractors \cite{ren2021knowledge}. The current state-of-the-art methods have shifted toward candidate generating-and-ranking (CGR) frameworks using pre-trained language models \cite{chiang2022cdgp} and transformer-based architectures that can capture contextual information \cite{wang2023distractor}. Recent innovations by \cite{taslimipoor2024distractor} have further demonstrated the effectiveness of combining generative and discriminative capabilities of transformer models to produce high-quality distractors across various question types.

When generating distractors automatically, models often produce mixed-quality candidates. As the example shown in Table \ref{tab:example}, for the question "If an object is attracted to a magnet, the object is most likely made of \_\_\_" (answer: "metal"), a model might generate both effective distractors like "wood" and "plastic" alongside problematic ones like "steel" and "iron" (which are types of metals). Rather than discarding imperfect candidates, we propose learning from this full spectrum by differentially rewarding distractors based on their quality. This approach helps the model distinguish between pedagogically effective options and those that are semantically related but contextually inappropriate for assessment purposes.

\begin{table}[htbp]
    \centering
    \begin{tabular}{l l}
        \toprule
        \multirow{2}{*}{\textbf{Cloze Test Sentence}} & If an object is attracted to a magnet,  \\
        & the object is most likely made of \_\_\_\_.\\
        \midrule
        \textbf{Correct Answer} & metal \\
        \textbf{Gold Standard Distractors} & wood, plastic, cardboard \\
        \textbf{Generated Distractors} & wood, plastic, glass, stone, steel, aluminum, clay, iron, rubber \\
        \bottomrule
    \end{tabular}
    \caption{Distractors Generation Example}
    \label{tab:example}
\end{table}

We introduce DualReward, a novel reinforcement learning framework for automatic distractor generation that addresses key limitations of existing approaches through two primary innovations. First, we propose a dual reward structure that differentiates between gold standard human-created distractors and model-generated candidates, enabling the model to learn from both reliable examples and its own generations while maintaining appropriate distinctions between them. Second, we develop an adaptive reward scaling mechanism that dynamically adjusts the intensity of learning signals based on model performance, providing stronger guidance when the model struggles with difficult examples and more subtle refinement as performance improves. 

Our approach draws conceptual inspiration from recent advances in reinforcement learning, particularly adaptive reward mechanisms \cite{haarnoja2018soft} and curriculum learning \cite{florensa2018automatic}, while applying these principles to the unique challenges of educational content generation. Unlike previous approaches that treat all distractors uniformly, our method acknowledges the inherent difference in reliability between human-created and model-generated options, applying differential rewards scaled by model confidence.

We evaluate our DualReward framework on two established benchmarks: CLOTH-F \cite{wang2023distractor}, a passage-level cloze dataset, and MCQ \cite{ren2021knowledge}, a diverse cross-domain sentence-level dataset. Our results demonstrate that the proposed approach consistently outperforms state-of-the-art baselines across multiple metrics, with particularly notable improvements in ranking quality as measured by Mean Reciprocal Rank (MRR) and Normalized Discounted Cumulative Gain (NDCG).

Our contributions can be summarized as follows:
\begin{itemize}
\item We introduce a dual reward structure for distractor generation that effectively differentiates between gold standard and model-generated distractors, enabling more nuanced learning from both sources.
\item We develop an adaptive reward scaling mechanism that dynamically adjusts learning signals based on model performance, facilitating more effective training without extensive hyperparameter tuning.
\item We demonstrate through comprehensive experiments that our DualReward framework achieves state-of-the-art performance on both passage-level and sentence-level cloze tests, with significant improvements in distractor ranking quality.
\end{itemize}

Our work bridges the gap between recent advances in reinforcement learning and automated educational assessment, offering a flexible framework that can be applied to diverse cloze test formats and domains. We provide our codes at https://github.com/tenny314159/dualreward.

\section{Related Research}
Our work on adaptive reward-scaled reinforcement learning for distractor generation builds upon two primary research areas: (1) automatic distractor generation for educational assessments and (2) adaptive reward mechanisms in reinforcement learning.

\subsection{Automatic Distractor Generation}
Automatic distractor generation (ADG) has become increasingly important in educational technology to reduce the manual effort in creating high-quality assessment items \cite{alhazmi2024distractor}. Several approaches have been proposed for multiple-choice questions, ranging from context-sensitive inference rules \cite{zesch2014distractor} to difficulty-aware generation \cite{yeung2019distractor}.

Knowledge-driven methods \cite{ren2021knowledge} leverage external knowledge bases to identify semantically related but incorrect options. Transformer-based approaches employ both generative and discriminative capabilities to produce plausible distractors \cite{taslimipoor2024distractor}. Chung et al. \shortcite{chung2020distractor} proposed a BERT-based distractor generation scheme with multi-tasking and negative answer training strategies.

For cloze tests specifically, several specialized methods have been developed. Chiang et al.\shortcite{chiang2022cdgp} introduced CDGP, an automatic cloze distractor generation approach based on pre-trained language models. Panda et al. \shortcite{panda2022distractor} explored round-trip neural machine translation for generating distractors in fill-in-the-blank exercises. Jiang \& Lee \shortcite{jiang2017distractor} focused on Chinese fill-in-the-blank items, while Zhang et al. \shortcite{zhang2023clozex} introduced ClozEx for generating explanations for English cloze tests.

Recent advancements include DiVERT \cite{fernandez2024divert}, which generates distractors with variational errors for mathematics questions, and approaches leveraging large language models with counterfactual contrastive decoding \cite{qu2024unsupervised} and knowledge graph integration \cite{yu2024enhancing}. Kumar et al. \shortcite{kumar2023novel} proposed a novel approach incorporating semantic similarity measures and contextual understanding to create effective distractors.

\subsection{Adaptive Reward Mechanisms in Reinforcement Learning}
Our reinforcement learning component builds on adaptive reward mechanisms developed in recent years. Haarnoja et al.\shortcite{haarnoja2018soft} introduced Soft Actor-Critic with automatically adjusted temperature parameters, conceptually related to our adaptive reward scaling. This approach optimizes entropy-regularized policies, balancing exploration and exploitation, similar to how our method balances confidence in generated distractors.

Policy gradient methods with dynamic reward scaling Schulman et al. \shortcite{schulman2017ppo} provide a foundation for our approach by demonstrating how adaptive scaling factors can stabilize training. Hadfield-Menell et al.\shortcite{hadfieldmenell2017inverse} explored inverse reward design, highlighting the importance of dynamic reward structures in complex learning environments. Their work on reward uncertainty parallels our confidence-weighted approach.

In practice-oriented research, Wang et al. \shortcite{wang2020perturbed} proposed methods for handling noisy rewards that share conceptual similarities with our approach to handling uncertain distractor quality. Florensa et al. \shortcite{florensa2018automatic} introduced reverse curriculum generation, where task difficulty progressively increases as the agent improves, conceptually similar to our adaptive scaling mechanism that adjusts based on model performance.

Kwon et al.\shortcite{kwon2024adaptive} presented adaptive reward design for reinforcement learning in complex robotic tasks with an emphasis on dynamic difficulty adjustment. In the context of curriculum learning, Narvekar et al. \shortcite{narvekar2020curriculum} reviewed approaches where task difficulty adapts based on learner progress, paralleling our method of adjusting reward scaling based on model performance.

Meta-gradient reinforcement learning \cite{xu2018metagradient} provides theoretical foundations for hyperparameter adaptation that inspired our sigmoid-based reward scaling mechanism. Similarly, Zheng \shortcite{zheng2018intrinsic} developed methods for learning intrinsic reward functions that can be dynamically adapted during training, which informed our approach to automatically adjusting reward scales based on training progress.

Our work bridges these research areas by introducing an adaptive reward-scaled reinforcement learning framework specifically designed for distractor generation. Unlike previous approaches that rely primarily on supervised learning or static reward structures, our method dynamically adjusts reward signals based on model performance and confidence, with a dual reward structure that differentiates between gold standard and model-generated distractors.

\section{Methodology}

\begin{figure*}
\centering
\includegraphics[scale=0.5]{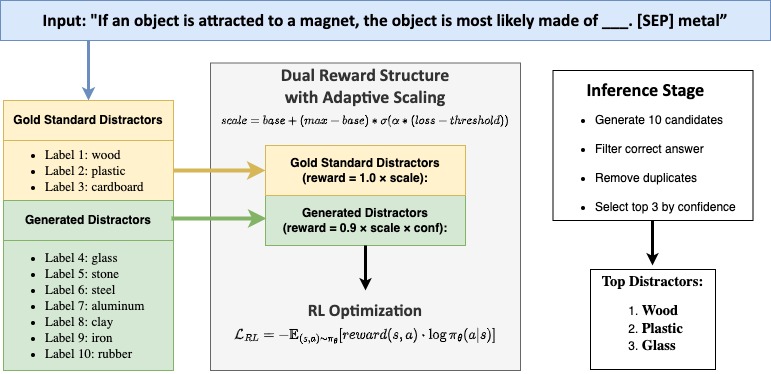}
\caption{DualReward Framework} 
\label{framework}
\end{figure*}

Our approach addresses automatic distractor generation for cloze tests using a adaptive reward-scaled reinforcement learning framework. Figure \ref{framework} provides an overview of our approach using a concrete example: for the cloze test ``If an object is attracted to a magnet, the object is most likely made of \_\_\_\_" with the correct answer ``metal", our framework processes this input through four key components: (1) a distractor candidate generation model that produces gold standard distractors (wood, plastic, cardboard) and model-generated candidates (glass, stone, steel, aluminum, clay, iron, rubber), (2) an adaptive reward scaling mechanism, (3) a dual reward structure that differentiates between these two types of distractors, and (4) an inference procedure that selects optimal distractors based on confidence scores. 

\subsection{Training Data Generation}

Our distractor generation process consists of the following steps:

\begin{itemize}
\item Initial Candidate Generation: For each cloze test instance, we generate an initial pool of candidate distractors using our base language model.

\item Candidate Filtering: We filter out candidates that are identical to any gold standard distractors from the training set, ensuring the model learns to generate novel distractors rather than memorizing examples.

\item Supplementary Generation: If fewer than 7 unique candidates remain after filtering, we perform additional generation passes until we obtain 7 distinct candidate distractors.

\item Training Data Preparation: For each training instance, we compile a set of 10 potential answers (labels): 3 gold standard distractors from human-created references and 7 model-generated candidate distractors.
\end{itemize}

We conducted pilot study to empirically test different pool sizes of candidate distractors, including 10 generated candidates + 3 gold standar (total 13). Results remained similar, so we chose 7 generated + 3 gold standard for a balanced total of 10 candidates. 

\subsection{Dual Reward Structure with Adaptive Scaling}
The core  of our approach lies in a dual reward structure with adaptive scaling that differentiates between gold standard distractors and model-generated candidates. Our framework draws conceptual inspiration from asymmetric learning paradigms in other domains. For instance, Zhang and Lapata  \shortcite{zhang2017simplification} used asymmetric rewards in sentence simplification. However, we design differential rewards between gold-standard and model-generated distractors. Human-created distractors are treated as fully reliable examples.  We use the Candidate Set Generation (CSG) module from the CDGP model by Chiang et al. \shortcite{chiang2022cdgp} to generate candidate distractor sets. In CDGP, each candidate distractor receives a $confidence\_score$ from BERT's masked language modeling head, computed as the softmax probability of the distractor token at the masked position. Model-generated distractors receive a discounted reward based on model confidence. The discount factor reflects inherent uncertainty in model-generated versus human-validated content. The 0.9 value provides meaningful but not excessive penalty, encouraging exploration while preventing overconfidence in generated distractors.
\\
\\
\noindent
\textbf{Gold Standard Distractors}: 
$$reward_{gold} = 1 \times reward\_scale$$
\textbf{Generated Candidate Distractors}: 
$$reward_{gen} = 0.9 \times reward\_scale \times confidence\_score$$
where $confidence\_score$ represents the model's confidence in the candidate distractor's suitability.
\\
\\
\noindent\textbf{Adaptive Reward Scaling}

Our adaptive reward scaling mechanism builds upon principles established in reinforcement learning literature. The sigmoid-based transition function resembles the temperature adaptation in Soft Actor-Critic \cite{haarnoja2018soft}, where the temperature parameter dynamically adjusts to balance exploration and exploitation. The specific form of our scaling equation:
$$reward\_scale = base\_scale + (max\_scale - base\_scale) \cdot \sigma(\alpha \cdot (avg\_loss - threshold))$$

where:
\begin{itemize}
\item $base\_scale$ represents the minimum reward scaling factor
\item $max\_scale$ represents the maximum reward scaling factor
\item $\alpha$ controls the steepness of the transition between scaling factors
\item $threshold$ defines the loss value at which the scaling begins to transition
\item $\sigma$ is the sigmoid function $\sigma(x) = \frac{1}{1+e^{-x}}$
\item $avg\_loss$ is the average loss during the current training phase
\end{itemize}

The dual rewards establish a hierarchy between gold-standard and generated distractors, while the adaptive scaling dynamically adjusts the overall intensity of learning signals based on model performance. This integrated approach allows the model to simultaneously learn from reliable examples while exploring the generation of novel distractors, with the learning signal strength automatically calibrated to the current stage of training.

We optimize the model using a reinforcement learning objective that incorporates these scaled rewards:

$$\mathcal{L}_{RL} = -\mathbb{E}_{(s,a) \sim \pi_\theta}[reward(s,a) \cdot \log \pi_\theta(a|s)]$$

where $\pi_\theta$ represents our policy (the distractor generation model), $s$ is the input sentence, and $a$ is the generated distractor.

\subsection{Inference Procedure}

During test time, we employ the following procedure to select the final distractors:
\begin{itemize}
\item Generate 10 candidate distractors for the given cloze test
\item Filter out candidates that are identical to the correct answer
\item Remove duplicate candidates
\item Select the top 3 candidates based on the model's confidence scores
\end{itemize}
This approach ensures that the final distractors are diverse, distinct from the correct answer, and optimize the model's learned criteria for effective distractors.

\section{Experiment Design}

\subsection{Dataset}
We evaluate our DualReward framework on two established benchmarks for cloze test distractor generation.  \textbf{CLOTH-F (Filtered)} is a refined subset of the CLOTH dataset, introduced by \cite{wang2023distractor}. The original CLOTH dataset contains questions with numbered blanks (e.g., "1"). To avoid training data inconsistency and create a more standardized evaluation setting, CLOTH-F removes the questions with numbered blanks, resulting in 5,041/720/739 instances for train/dev/test splits. Each instance consists of a passage-level cloze test with a blank indicated by an underscore, a correct answer, and human-generated distractors. 

\textbf{MCQ} is a cross-domain, sentence-level cloze dataset spanning science, vocabulary, common sense, and trivia domains. It contains 2,321/258 questions for train/test splits (with dev data created by a 9:1 split from training), where each instance has a sentence-level context with blanks marked as blank. While CLOTH-F provides rich contextual information through full passages, MCQ offers domain diversity with more focused sentence-level questions, allowing us to evaluate our model's performance across different cloze test formats and difficulty levels. Table \ref{tab:statistics} provides the dataset statistics.

\begin{table}[htbp]
    \centering
    \begin{tabular}{lrrrrrrrr}
        \toprule
        \multirow{2}{*}{Dataset} & \multicolumn{4}{c}{CLOTH-F} & \multicolumn{4}{c}{MCQ} \\
        \cmidrule(lr){2-5} \cmidrule(lr){6-9}
                              & Train  & Dev   & Test  & All   & Train & Dev   & Test  & All   \\
        \midrule
        \# of Passages         & 5,041  & 720   & 739   & 6500  & -     & -     & -     & -     \\
        \# of Questions        & 69,009 & 9,696 &10,233 &88,938 & 2088  & 233   & 258   &2580   \\
        \bottomrule
    \end{tabular}
     \caption{Statistics for the training, development, and test sets of CLOTH-F and MCQ. }
    \label{tab:statistics}
\end{table}

\subsection{Baseline Models}

We compare with the following baselines on experiments with CLOTH-F dataset:

\textbf{T5 multi-task (+ CTA)}: \cite{wang2023distractor} enhance T5 distractor generation through multi-task training with cloze test answering (CTA). The model jointly learns to generate distractors and answer cloze questions, where CTA requires selecting the correct answer from shuffled options. This dual-task approach improves the model's understanding of answer-distractor relationships.

\textbf{T5 multi-task (+ DF, CTA)}: \cite{wang2023distractor} extend their multi-task framework by adding distractor finding (DF) to CTA. The DF task trains the model to identify distractors placed in cloze gaps, creating a three-task training setup with balanced data sampling. This comprehensive approach achieves their best results on CLOTH-F with 28.47 P@1.

We compare with the following baselines on experiments with MCQ dataset:

\textbf{T5 multi-task (+DF)}: \cite{wang2023distractor} enhances T5 distractor generation through multi-task training with the distractor finding (DF) task. The DF task trains the model to identify distractor spans within cloze passages by placing a distractor at the blank position and having the model generate that distractor based on the modified input. This additional task helps the model better understand distractor characteristics.

\textbf{CSG-DS}: \cite{ren2021knowledge} employs a candidate generating-and-ranking framework consisting of two stages: a Candidate Set Generator (CSG) that extracts semantically similar candidates from knowledge bases using context-dependent conceptualization, and a Distractor Selector (DS) that re-ranks candidates using a learning-to-rank model with elaborately designed features measuring both plausibility and reliability.

\textbf{CDGP} \cite{chiang2022cdgp} leverages pre-trained language models (specifically BERT) for distractor generation within the candidate generating-and-ranking paradigm. The model fine-tunes BERT to predict distractors from the masked blank position, using the language model's inherent fill-in-the-blank capability from masked language modeling pretraining.

\textbf{RAP-Pt5}: \cite{yu2024enhancing} introduces Retrieval Augmented Pretraining for the DG task. This approach uses MCQ answers to retrieve relevant sentences/passages from large corpora like Wikipedia to create pseudo questions for task-specific pretraining. In this specific implementation, the T5 model is first pretrained on the Sciq-all dataset using RAP before being fine-tuned for MCQ distractor generation, providing cross-domain knowledge transfer from the larger Sciq dataset to improve MCQ performance.

\subsection{Implementation Details}
For the initial candidate generation phase, we apply BERT within the Candidate Set Generation (CSG) framework established by \cite{chiang2022cdgp}. We employ T5-base as our foundational language model for distractor selection experiments. When conducting experiments on the MCQ dataset, we first pretrain the T5-base model on the Sciq-all dataset before implementing our DualReward framework, enabling cross-domain knowledge transfer from the larger scientific corpus following \cite{yu2024enhancing}.

Our hyperparameter configuration includes 20 training epochs with a batch size of 8, utilizing a learning rate of 1e-4 with the AdamW optimizer. Our adaptive scaling provides stronger reinforcement signals when the model struggles (high loss) and more subtle guidance as performance improves. We set $base\_scale = 0.1$ and $max\_scale = 0.2$ to provide a controlled range of reward magnitudes, following similar bounded scaling approaches in policy gradient methods \cite{schulman2017ppo}. The steepness parameter $\alpha = 5$ was selected to create a responsive yet stable transition, similar to sensitivity parameters in curriculum learning \cite{florensa2018automatic}. The threshold = 1.0 was determined based on preliminary experiments, representing an inflection point in learning dynamics where model confidence begins to stabilize.

For comprehensive evaluation, we adopt the standard metrics established by state-of-the-art models in distractor generation. These include Precision@1 (P@1), which measures the accuracy of the top-ranked distractor; Recall@1 (R@1), which evaluates the proportion of ground-truth distractors retrieved at rank one; F1@3, representing the harmonic mean of precision and recall for the top three predictions; Mean Reciprocal Rank@3 (MRR@3), which captures the ranking quality by computing the reciprocal of the rank at which the first correct distractor appears; and Normalized Discounted Cumulative Gain@3 (NDCG@3), which provides a graded relevance assessment with position-based discounting. Together, these metrics offer a comprehensive view of both generation quality and ranking effectiveness.

\section{Results and Analysis}
\subsection{Main Results}
The experimental results in Table \ref{tab:main_results} demonstrate the effectiveness of our proposed DualReward framework across both datasets. On the CLOTH-F dataset, our approach achieves state-of-the-art performance with a P@1 of 28.87\%, slightly outperforming the strong T5 multi-task baseline methods. While the absolute improvement in precision metrics is modest (0.4\% over the CTA baseline), our method shows more substantial gains in ranking quality metrics, with a 2.06\% improvement in MRR@3 and a remarkable 6.56\% gain in NDCG@3 compared to the best baseline.

These improvements in ranking metrics are particularly significant as they indicate that our method not only identifies the most plausible distractor correctly but also produces a more effective overall ranking of distractor candidates. The F1@3 score of 21.87\% represents a 2.05\% absolute improvement over the best baseline, demonstrating better precision-recall balance across the top three predictions.

On the MCQ dataset, the advantages of our approach are even more pronounced. The DualReward framework with pretraining achieves 37.84\% P@1, surpassing the strong RAP-PT5 baseline by 6.18\%. This substantial improvement is consistent across all metrics, with particularly notable gains in MRR@3 (8.5\% improvement) and NDCG@3 (14.23\% improvement). These results highlight that our adaptive reward scaling mechanism is especially effective in the cross-domain setting with sentence-level cloze tests, where the model must handle diverse topics and question formats.

\begin{table}[htbp]
    \centering
    \begin{tabular}{l l c c c c c}
        \toprule
        \multicolumn{1}{l}{Datasets} & \multicolumn{1}{l}{Model} & P@1 & R@1 & F1@3 & MRR@3 & NDCG@3 \\
        \midrule
        \multirow{3}{*}{CLOTH-F} & T5 multi-task (+ CTA) & 28.75 & 9.58 & 19.20 & 34.06 & 35.64 \\
        & T5 multi-task (+ DF, CTA) & 28.47 & 9.49 & 19.82 & 34.46 & 36.26 \\
        & Dual-Reward & \textbf{28.87} & \textbf{9.63} & \textbf{21.87} & \textbf{36.52} & \textbf{42.82} \\
        \midrule
        \multirow{4}{*}{MCQ} & T5 multi-task (+ DF) & 22.00 & 7.33 & 13.64 & 27.15 & 28.50 \\
        & CSG+DS & 10.08 & - & 9.42 & 17.33 & - \\
        & CDGP & 13.13 & - & 12.23 & - & - \\
        & RAP-PT5 (w/ pretrain) & 31.66 & 10.55 & 18.91 & 37.70 & 39.43 \\
        & Dual-Reward (w/pretrain) & \textbf{37.84} & \textbf{12.61} & \textbf{23.42} & \textbf{46.20} & \textbf{53.66} \\
        \bottomrule
    \end{tabular}
    \caption{Main Results}
    \label{tab:main_results}
\end{table}

\subsection{Reward Scale}
\begin{figure*}
\centering
\includegraphics[scale=0.04]{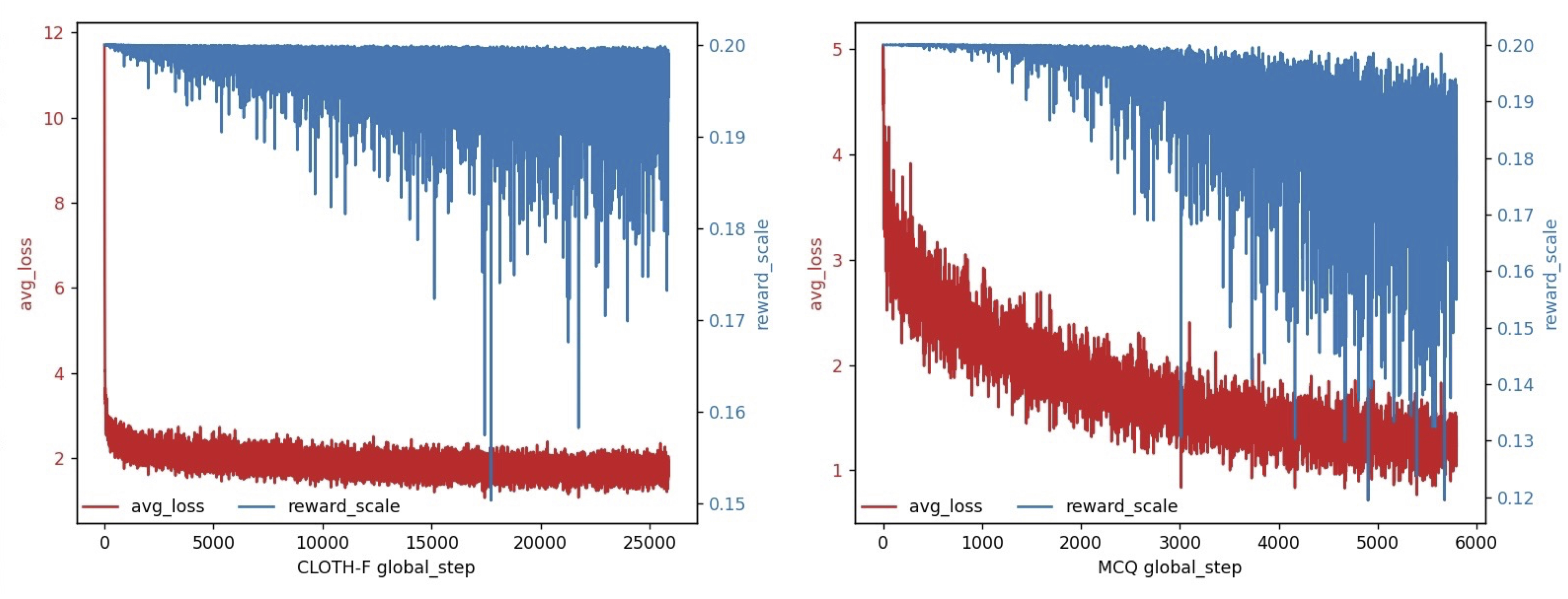}
\caption{Change of Reward Scale} 
\label{rewardScale}
\end{figure*}

Figure \ref{rewardScale} illustrates the dynamic nature of our adaptive reward scaling mechanism during training on both CLOTH-F and MCQ datasets. As shown in the figure, the reward scale (right y-axis, blue line) changes in response to the average loss (left y-axis, red line) throughout the training process. For the CLOTH-F dataset, we observe that the reward scale starts at a higher value and gradually decreases as the model's performance improves and the loss decreases. This pattern demonstrates how our mechanism provides stronger learning signals during the initial training phases when the model is less confident, and then transitions to more subtle guidance as performance stabilizes. The MCQ dataset shows a similar but more volatile pattern, with the reward scale fluctuating more frequently in response to the diverse, cross-domain nature of the data. These visualizations support our quantitative findings, confirming that adaptive scaling effectively responds to the model's changing learning dynamics, particularly when handling varied question types and domains.

\subsection{Ablation Study}

\subsubsection{Dual Reward vs. Uniform Reward}
To further validate the importance of our adaptive reward scaling mechanism, we conducted an ablation study where we replaced the dynamic reward scale with constant values. This differs from our dual reward approach where model-generated distractors receive a discounted reward ($0.9 \times reward\_scale \times confidence\_score$).

$$reward_{gold} = 1 \times reward\_scale$$
$$reward_{gen} = 1 \times reward\_scale \times confidence\_score$$

As shown in Table \ref{tab:ablation_study}, the dual reward structure consistently outperforms the uniform reward approach. On CLOTH-F, our method achieves a 0.53\% improvement in P@1 (28.87\% vs. 28.34\%) and more substantial gains in ranking metrics (0.66\% in MRR@3, 0.71\% in NDCG@3). For the MCQ dataset, the improvements are more pronounced, with a 4.64\% gain in P@1 (37.84\% vs. 33.20\%) and similar improvements in ranking metrics.
These results confirm that establishing a clear hierarchy between human-created examples and model-generated candidates enables more effective learning, particularly in diverse, cross-domain settings like the MCQ dataset.

\subsubsection{Adaptive Reward vs. Constant Reward}
To evaluate our adaptive reward scaling mechanism, we compared it against three fixed reward scale values (0.1, 0.15, and 0.2) while maintaining the dual reward structure. This eliminates the adaptive component that dynamically adjusts based on model performance during training:

$$reward_{gold} = 1 \times constant\_reward$$
$$reward_{gen} = 0.9 \times constant\_reward \times confidence\_score$$

As shown in Table \ref{tab:ablation_study}, the adaptive approach consistently outperforms fixed scaling across all metrics on both datasets. On CLOTH-F, the improvements are modest but consistent, with gains of 0.21-0.46\% in P@1 over the constant reward baselines. The MCQ dataset shows more substantial benefits from adaptive scaling, with improvements of 3.48-3.86\% in P@1 compared to the constant reward variants.

The more pronounced improvements on MCQ suggest that adaptive scaling is particularly beneficial for diverse, cross-domain datasets where learning dynamics may vary more substantially during training. When encountering questions from unfamiliar domains, the adaptive mechanism provides stronger learning signals, facilitating more effective transfer learning.

Analysis of the results across constant reward settings reveals that no single fixed value works optimally across different datasets or metrics, underscoring the value of our adaptive approach, which automatically calibrates reward intensity based on the model's current learning state without requiring extensive hyperparameter tuning. We also performed an ablation study on the MCQ dataset where we did not pretrain T5-base model on the Sciq-all dataset before implementing our DualReward framework. The results as shown in the last row of Table 4 are lower than the Dual-Reward with pretrain.

\begin{table}[htbp]
    \centering
    \begin{tabular}{l l c c c c c}
        \toprule
        \multicolumn{1}{l}{Datasets} & \multicolumn{1}{l}{Model} & P@1 & R@1 & F1@3 & MRR@3 & NDCG@3 \\
        \midrule
        \multirow{4}{*}{CLOTH-F} & Dual-Reward & \textbf{28.87} & \textbf{9.63} & \textbf{21.87} & \textbf{36.52} & \textbf{42.82} \\
        & w/ constant reward = 0.1 & 28.66 & 9.55 & 21.55 & 36.09 & 42.21 \\
        & w/ constant reward = 0.15 & 28.41 & 9.47 & 21.65 & 36.01 & 42.31 \\
        & w/ constant reward = 0.2 & 28.56 & 9.52 & 21.69 & 36.21 & 42.58 \\
        &  Uniform-Reward & 28.34 & 9.45 & 21.38 & 35.86 & 42.11 \\
        \midrule
        \multirow{4}{*}{MCQ} & Dual-Reward (w/pretrain) & \textbf{37.84} & \textbf{12.61} & \textbf{23.42} & \textbf{46.20} & \textbf{53.66} \\
        & w/ constant reward = 0.1 & 33.98 & 11.33 & 23.17 & 42.92 & 50.49 \\
        & w/ constant reward = 0.15 & 34.36 & 11.45 & 23.29 & 43.82 & 52.19 \\
        & w/ constant reward = 0.2 & 33.98 & 11.33 & 21.88 & 42.34 & 49.59 \\
        &  Uniform-Reward & 33.20 & 11.07 & 23.42 & 42.54 & 50.75 \\
        & Dual-Reward (w/o pretrain) & 22.39 & 7.46 & 13.64 & 27.86 & 32.53\\
        \bottomrule
    \end{tabular}
    \caption{Ablation Study of Adaptive Reward Scale and Uniform Reward}
    \label{tab:ablation_study}
\end{table}

\subsection{Qualitative Analysis}
To gain deeper insights into the effectiveness of our DualReward framework, we present sample outputs in Tables 5 and 6 for MCQ and CLOTH-F datasets, respectively. These examples demonstrate how our approach improves distractor quality compared to the Uniform-Reward baseline.

In the MCQ examples (Table \ref{mcq_example} ), our DualReward framework generates more contextually appropriate distractors. For instance, in Sentence 1 regarding transverse waves, our model produces "peaks, ridges, valleys" versus the baseline's "plateaus, peaks, valleys," where "plateaus" is less appropriate for wave physics terminology. In Sentence 4 about photosynthesis, our model generates "carbon, helium, hydrogen" while the baseline includes "water," which is problematic since water is actually a photosynthesis reactant.

For the CLOTH-F passage (Table \ref{cloth_example}), our framework demonstrates better contextual awareness. In blank 1 where the answer is "eating," our model generates "working, studying, thinking" - all plausible breakfast meeting activities, while the baseline produces "sleeping, playing, studying" where "sleeping" is contextually inappropriate. Similarly, for blank 11 (answer: "decision"), our model generates "plan, promise, suggestion" - all commitment-related terms, versus the baseline's inclusion of "explanation" which fits less well in the narrative context.

While our DualReward framework generally produces better distractors, some limitations remain. In MCQ Sentence 7, our model generates "leaves, stems, seeds" for gymnosperms that "do not have flowers," where "seeds" is problematic since gymnosperms actually do have seeds. Similarly, in CLOTH-F blank 10 (answer: "hit"), our model produces "beat, pleased, attracted" where "beat" could be considered too close in meaning to the correct answer.

These examples demonstrate that while our DualReward framework generally improves distractor quality through better contextual appropriateness, challenges remain in avoiding semantic overlaps and factual inconsistencies.

\section{Conclusion}
In this paper, we introduced DualReward, a novel reinforcement learning framework for automatic cloze test distractor generation that employs both adaptive reward scaling and a dual reward structure. Our approach dynamically adjusts learning signals based on model performance while differentiating between gold standard and model-generated distractors, addressing key limitations in existing supervised learning approaches. Experimental results on both CLOTH-F and MCQ benchmarks demonstrate consistent improvements over state-of-the-art approaches, with particularly notable gains on the diverse MCQ dataset (6.18\% improvement in P@1, 8.5\% in MRR@3, and 14.23\% in NDCG@3). Comprehensive ablation studies confirm the benefits of our adaptive reward scaling mechanism, especially in cross-domain settings where learning dynamics vary substantially during training.

While our evaluation primarily focuses on matching human-created distractors rather than directly measuring pedagogical effectiveness, the DualReward framework provides a promising foundation for balancing learning from human-created examples while generating novel distractors. Future research directions include extending the framework to multilingual cloze tests, incorporating pedagogically-aligned evaluation metrics that measure actual learning outcomes, and exploring applications to other educational assessment formats such as reading comprehension and mathematical problem generation. The adaptive reward mechanism also holds potential for broader applications in other text generation tasks requiring quality differentiation between training examples.

\section{Limitations}
Despite promising results, our approach has several limitations. First, our evaluation focuses on matching human-created distractors rather than measuring actual educational effectiveness, leaving pedagogical impact incompletely assessed. Second, experiments are limited to English cloze tests due to the absence of established cross-lingual benchmarks in this domain. Future work should incorporate pedagogically-aligned evaluation metrics that measure learning outcomes and extend to multilingual settings as appropriate datasets become available.


\newpage
\section{Appendix}

\begin{table}[htbp]
    \centering
	\setlength{\extrarowheight}{-2pt} 
    \begin{tabular}{l l}
		\toprule
        \textbf{Sentence} & 1. The high points of a transverse wave are called \_\_\_ . \\
        \textbf{Answer / Distractor} & crests / peaks, troughs, ridges \\
        \textbf{Dual-Reward} & peaks, ridges, valleys \\
        \textbf{Uniform-Reward} & plateaus, peaks, valleys \\
        
        \midrule
        \textbf{Sentence} & 2. In sexual reproduction , \_\_\_ is the name of the gamete cell the male must contribute. \\
        \textbf{Answer / Distractor} & sperm / plasma, spore, ova \\
        \textbf{Dual-Reward} & egg, dna, fetus \\
        \textbf{Uniform-Reward} & egg, dna, rna \\
        
        \midrule
        \textbf{Sentence} & 3. The main organs of the respiratory system are \_\_\_ . \\
        \textbf{Answer / Distractor} & lungs / ovaries, intestines, kidneys \\
        \textbf{Dual-Reward} & ovaries, kidneys, liver \\
        \textbf{Uniform-Reward} & ovaries, kidney, liver \\
        
        \midrule
        \textbf{Sentence} & 4. The products of photosynthesis are glucose and \_\_\_ else . \\
        \textbf{Answer / Distractor} & oxygen / carbon, hydrogen, nitrogen \\
        \textbf{Dual-Reward} & carbon, helium, hydrogen \\
        \textbf{Uniform-Reward} & nitrogen, hydrogen, water \\
        
        \midrule
        \textbf{Sentence} & 5. \_\_\_ is responsible for erosion by flowing water and glaciers . \\
        \textbf{Answer / Distractor} & gravity / kinetic, electromagnetic, weight \\
        \textbf{Dual-Reward} & weight, magnetism, electricity \\
        \textbf{Uniform-Reward} & motion, friction, magnetism\\
        
        \midrule
        \textbf{Sentence} & 6. \_\_\_ of rocks form from cooled magma or lava . \\
        \textbf{Answer / Distractor} & igneous / granite, sedimentary, metamorphic \\
        \textbf{Dual-Reward} & metamorphic, sedimentary, meta \\
        \textbf{Uniform-Reward} & metamorphic, granite, lava \\
        
        \midrule
        \textbf{Sentence} & 7. Gymnosperms have seeds but do not have \_\_\_ . \\
        \textbf{Answer / Distractor} & flowers / leaves, stems, roots \\
        \textbf{Dual-Reward} & leaves, stems, seeds \\
        \textbf{Uniform-Reward} & leaves, stems, stem \\
        
        \midrule
        \textbf{Sentence} & 8. Some 96 \% of the dry mass consists of organic compounds produced by \_\_\_ . \\
        \textbf{Answer / Distractor} & photosynthesis / erosion, electrolysis, reproduction \\
        \textbf{Dual-Reward} & spermatogenesis, glycolysis, erosion \\
        \textbf{Uniform-Reward} & glycolysis, spermatogenesis, osmosis \\
        
        \midrule
        \textbf{Sentence} & 9. It is called \_\_\_ when something is unable to move from place to place . \\
        \textbf{Answer / Distractor} & sessile / Undrate, obovate, ovate \\
        \textbf{Dual-Reward} & elongated, spile, ovate \\
        \textbf{Uniform-Reward} & spongile, elliptical, aqueous \\
        
        \midrule
        \textbf{Sentence} & 10. The greatest contribution of arthropods to human food supply is \_\_\_ . \\
        \textbf{Answer / Distractor} & pollination / vegetation, reproduction, hibernation \\
        \textbf{Dual-Reward} & migration, precipitation, vegetation \\
        \textbf{Uniform-Reward} & migration, sedimentation, eva \\
        
        \bottomrule
    \end{tabular}
    \caption{MCQ Generated Distractors Example}
    \label{mcq_example}
\end{table}

\begin{table}[htbp]
    \centering
    \setlength{\tabcolsep}{3pt} 
    \begin{tabularx}{\textwidth}{X}
        \toprule
        \multicolumn{1}{c}{\textbf{Passage}}\\
        \midrule
        Years ago, a critical event occurred in my life that would change it forever. I met Kurt Kampmeir of Success Motivation Incorporation for breakfast. While we were  \underline{1}  , Kurt asked me, " John, what is your  \underline{2}  for personal growth?" 
        Never at a loss for words, I tried to find things in my life that might  \underline{3}  for growth. I told him about the many activities in which I was  \underline{4}  . And I went into a  \underline{5}  about how hard I worked and the gains I was making. I must have talked for ten minutes. Kurt  \underline{6}  patiently, but then he  \underline{7}  smiled and said, "You don't have a personal plan for growth, do you?" 
        "No, I  \underline{8}  ." "You know," Kurt said simply, "growth is not a(n)  \underline{9}  process." 
        And that's when it  \underline{10}  me. I wasn't doing anything  on purpose  to make myself better. And at that moment, I made the  \underline{11}  : I will develop and follow a personal growth plan for my  \underline{12}  . 
        That night, I talked to my wife about my  \underline{13}  with Kurt and what I had learned. 
        I  \underline{14}  her the workbook and tapes Kurt was selling. We  \underline{15}  that Kurt wasn't just trying to make a sale. He was offering a  \underline{16}  for us to change our lives and achieve our dreams. Several important things happened that day. First, we decided to  \underline{17}  the resources. But more importantly, we made a commitment to  \underline{18}  together as a couple. From that day on, we learned together, traveled together, and sacrificed together. It was a  \underline{19}  decision. While too many couples grow apart, we were growing together. \\
        \bottomrule
    \end{tabularx}
\end{table}
\vspace{-23pt} 
\begin{table}[htbp]
    \centering
    \setlength{\tabcolsep}{3pt} 
    \begin{tabularx}{\textwidth}{XXX}
        \toprule
        \textbf{Answer / Distractor} & \textbf{Dual-Reward} & \textbf{Uniform-Reward} \\
        \midrule
        1. eating / working, preparing, thinking & 1. working, studying, thinking & 1. sleeping, playing, studying \\
        2. plan / suggestion, demand, request & 2. way, method, idea & 2. dream, idea, way \\
        3. qualify / appeal, look, call & 3. appeal, provide, care & 3. arrange, provide, care \\
        4. involved / trapped, lost, bathed & 4. occupied, attracted, invited & 4. occupied, lost, included \\
        5. speech / lecture, discussion, debate & 5. lecture, discussion, report & 5. discussion, joke, story \\
        6. listened / calculated, drank, explained & 6. cried, laughed, drank & 6. cried, shouted, talked \\
        7. finally / eagerly, gradually, gratefully & 7. suddenly, luckily, immediately & 7. suddenly, luckily, immediately \\
        8. admitted / interrupted, apologized, complained & 8. refused, agreed, argued & 8. refused, agreed, argued \\
        9. automatic / slow, independent, changing & 9. natural, important, difficult & 9. natural, ordinary, important \\
        10. hit / confused, informed, pleased & 10. beat, pleased, attracted & 10. beat, attracted, comfort \\
        11. decision/comment, announcement, arrangement & 11. plan, promise, suggestion & 11. promise, suggestion, explanation \\
        12. life / progress, performance, investment & 12. progress, study, family & 12. family, career, study \\
        13. conversation / contract, negotiation, argument & 13. decision, competition, agreement & 13. connection, competition, agreement \\
        14. showed / lent, sold, offered & 14. gave, lent, sent & 14. sent, lent, brought \\
        15. recognized / recalled, defined,  declared & 15. believed, hoped, imagined & 15. hoped, imagined, guessed \\
        16. way / tool, method, rule & 16. method, chance, road & 16. gift, road, place \\
        17. buy / provide, give, deliver & 17. save, use, provide & 17. save, use, borrow \\
        18. grow / survive, move, gather & 18. work, study, gather & 18. live, work, study \\
        19. wise / difficult, random, firm & 19. difficult, simple, common & 19. difficult, simple, strange \\
        \bottomrule
    \end{tabularx}
    \caption{CLOTH - F Generated Distractors Example}
    \label{cloth_example}
\end{table}

\end{document}